# Novel approach to locate region of interest in mammograms for Breast cancer

BV Divyashree[1], Amarnath R[1], Naveen M[1], G Hemantha Kumar*[1]

*Abstract:* Locating region of interest for breast cancer masses in the mammographic image is a challenging problem in medical image processing. In this research work, the keen idea is to efficiently extract suspected mass region for further examination. In particular to this fact breast boundary segmentation on sliced rgb image using modified intensity based approach followed by quad tree based division to spot out suspicious area are proposed in the paper. To evaluate the performance DDSM standard dataset are experimented and achieved acceptable accuracy.

## 1. Introduction

Potentially fatal disease among women who has crossed 40 years is none other than breast cancer. The cells in the breast rampant their growth and form abnormal shapes of breast tissues leading to cancer .The cause for the disease is not known exactly. The disease has killed many of the women's life when they are not diagnosed in the early stage. Hence, early detection is the key to reduce the death rates from breast cancer and to increase the life span of a patient. Clinically advised successful tool for the early detection that discloses the cancer tissues in the breast up to two years before a patient or a physician can feel or see some symptoms in the breast is mammography [1, 2].

American cancer society estimates that approximately 252,710 women and 2,470 men are diagnosed as new cases of breast cancer. Also tells that among new cases the mortality rate is about 40,610 women and 460 men [3].

Masses, calcifications, architectural distortion and bilateral asymmetry observed are considered as abnormalities in the mammographic images. Also masses occur in different shapes like round, oval, speculated, nodular lobulated and stellate. However in the present work we are concentrating on finding the suspicious region for obscured masses which is very important action to root out breast cancer in the early stage [4,5]. Though mammograms are the efficient tool for early detection, there are many challenges in detecting mammographic lesions because all masses may not be cancer. The closely compacted tissues can hide some of the cancerous masses also, both looks white and in contrast, fatty tissue looks almost black on black background. Thus, pinpointing the region for the detection of cancer in the early stage to aid the radiologists survived as still hot task in automation. In the proposed method we have developed an efficient model to overcome the difficulties discussed.

In the proposed work firstly, RGB color bands are segregated separately, and layered each of them as multiple segments based on threshold. Secondly, each segment is analyzed for understanding the foreground (breast region) and background (background) region using masking technique. Thirdly, we performed intersection operation on layers of all channel. Then entropy is calculated on the segmented region using quad tree division to find the region of interest.

## 2. Background

Many researchers had proposed several techniques for the automation of breast cancer detection. Though it emerged as a challenging issue because the cancerous mass are subtle, infiltrative, looks almost like a normal tissue inside the dense scattered breast tissues and which is a place where radiologist's attention is needed. In the year 2006, Kolahdoorzan et al [8] proposed a breast pectoral muscle segmentation on digital mammograms using dyadic wavelet transform and by approximate bordering the breast. An approach was done by Houjinchen et al [9] to remove background from foreground by minimum cross entropy threshold method and to segment breast pectoral muscle using spectral clustering method. Another approach by Sreedevi and Sherly [10] proposed to remove breast pectoral muscle using global threshold, gray level threshold and canny edge detection along with a DCT based non local mean filter technique for preprocessing. In 2014, an attempt was made by Pereira et al [11] for segmentation and detection wherein they have used morphological operation followed by Ostu's thresholding method for segmentation and genetic algorithm for detection. Anuradha et al [12] proposed for segmentation, ROI extraction, watershed methods were proposed, which is compared with graph based saliency map with optimal threshold and regional maxima of saliency for accuracy. The saliency thresholding method attempted in ROI segmentation by not removing the pectoral muscle but the accuracy varies for chosen threshold on different images and also the morphological method depends on the structuring element. In 2015, Chun-Chu Jen and Shyr-Shen Yu [1] introduced a method for the detection of abnormal parts in mammograms where in segmentation is done using gray level quantization. Then, in the year 2016, Khalid El Fahssi et al [5] proposed a novel approach to classify abnormalities in the mammogram images. They have used shrink wrap function for preprocessing and for segmentation they have used combined approaches like level set theory based method and active contours. Combined top-hat and region growing methods for segmentation, classification by artificial neural networks were used by Karthikeyan et al [7]. Then algorithm for segmentation, based on intensity value, which are discontinuous, are also proposed by Jasmeen Kaur and MandeepKaur [13]. In 2017 by Luis Antonio Salazar-Licea et al [15] proposed a method for locating ROI using combined Shi-Tomasi corner detection, image thresholding and SIFT descriptors to improve the accuracy rates. However the literature presented provides some solution for the detection of breast cancer, yet it is still a challenging problem because the detection completely dependent on locating doubtful regions without missing any information. Segmenting foreground accurately with a single threshold value for different images is a challenging task, as it demands different threshold value for different images. Our motivation is that to provide an absolute technique for both (segmentation and ROI).

[1] University of Mysore, Department of studies in Computer Science , Manasagangotri, Mysore - 570006, India
* Corresponding Author Email: hemanthakumar@uni-mysore.ac.in

## 3. Proposed Algorithm

It is important not to lose any information from mammographic images, which may lead to misclassification of the malignant cases. In the proposed section, we introduced a new method for segmentation as input image. In the first stage we perform breast segmentation which are discussed in detail in coming section.

### 3.1. Segmentation

For segmenting, the mammographic images are taken as input. The image is split separately into red channel, green channel and blue channel. Then the separated channels are again sliced into five layers by fixing a threshold (Fig 1). This can be computed by the formula given Eq.1.

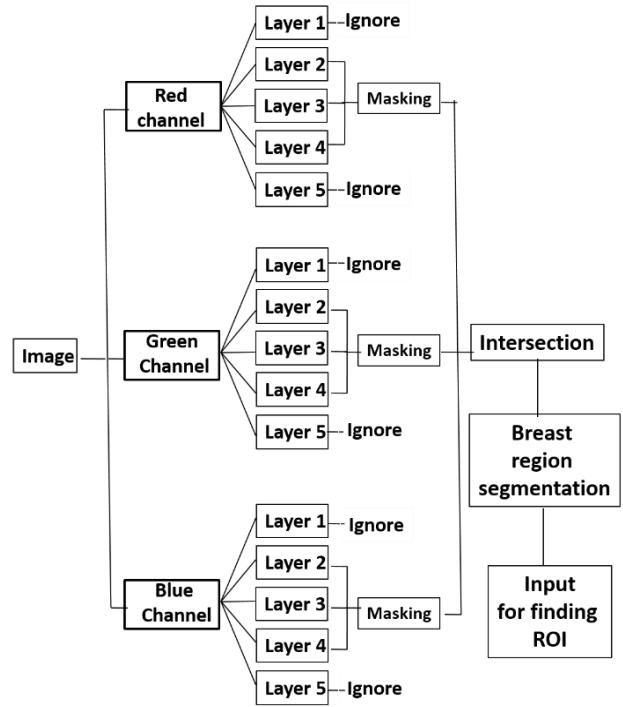

**Fig. 1.** Block diagram of segmentation

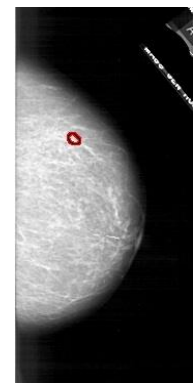

(a) Original image

$$\text{Red} = \begin{cases} \text{Layer1} = \text{Red}(i,j) < 50 \\ \text{Layer2} = \text{Red}(i,j) < 100 \\ \text{Layer2} = \text{Red}(i,j) < 150 \\ \text{Layer2} = \text{Red}(i,j) < 200 \\ \text{Layer2} = \text{Red}(i,j) < 255 \end{cases}$$

$$\text{Green} = \begin{cases} \text{Layer1} = \text{Green}(i,j) < 50 \\ \text{Layer2} = \text{Green}(i,j) < 100 \\ \text{Layer2} = \text{Green}(i,j) < 150 \\ \text{Layer2} = \text{Green}(i,j) < 200 \\ \text{Layer2} = \text{Green}(i,j) < 255 \end{cases} \quad (1)$$

$$\text{Blue} = \begin{cases} \text{Layer1} = \text{Blue}(i,j) < 50 \\ \text{Layer2} = \text{Blue}(i,j) < 100 \\ \text{Layer2} = \text{Blue}(i,j) < 150 \\ \text{Layer2} = \text{Blue}(i,j) < 200 \\ \text{Layer2} = \text{Blue}(i,j) < 255 \end{cases}$$

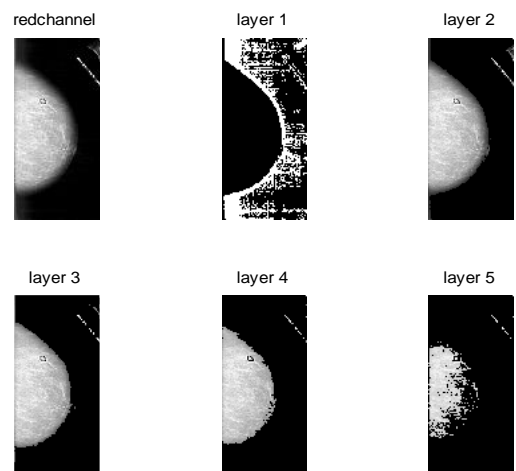

(b) Red channel and layers 1, 2, 3, 4, 5

First and the fifth layers of each channels does not provide required information (Fig. 2). So, rest of the layers of all the channels are used for segmentation purpose. The considered layers of the channels are again chopped up into number of blocks. The pixels in the blocks are separated into foreground and background regions based on the threshold value using masking operation.

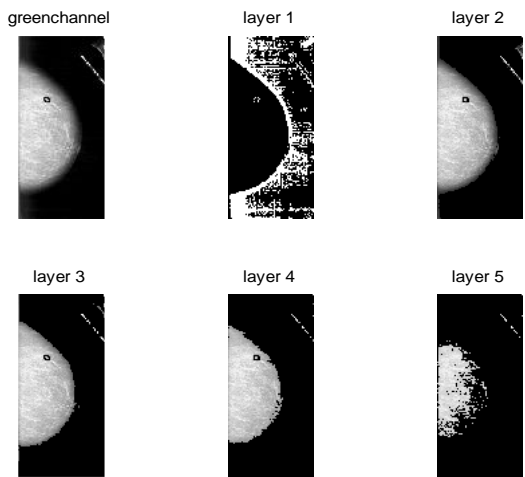

(c) Green channel and layers 1, 2, 3, 4, 5

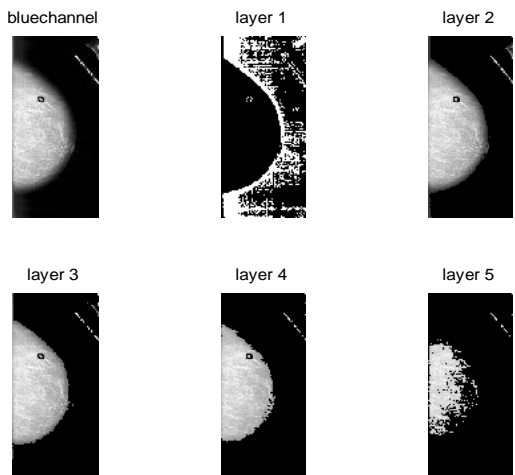

(d) Blue channel and **layers 1, 2, 3, 4, 5**

**Fig. 2.** Channel and Layers

Fig. 3. shows the segmented images of each channel layers wherein thresholding method made all pixels black and white considering threshold value.

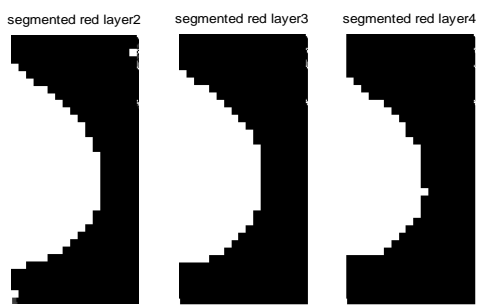

(a) Segmented red channel layers

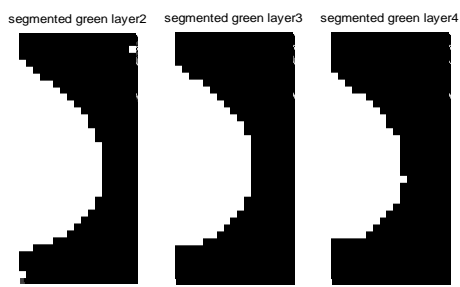

(b) Segmented green channel layers

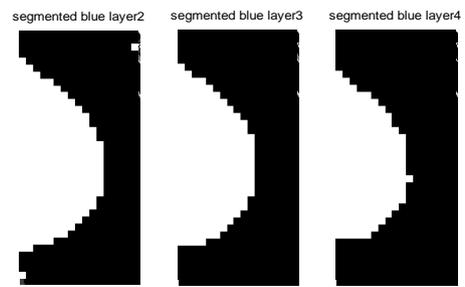

(c) Segmented blue channel layers

**Fig. 3.** Segmented channel and layers

**Algorithm for separating foreground from background**
1. Input image.
2. Take out red channel, green channel and blue channel.
3. Take red channel.
4. Initialize the size of the channel to height and width of the channel.
5. For j=1 and for i=1, find If channel(j,i)>=0 and channel(j,i)<=50, Then layer1=channel(j,i) else layer1=0.
6. Repeat step5 till j=height and i=width.
7. Endof the loop.
8. For j=1 and for i=1, find If channel(j,i)>50, Then layer2=channel(j,i) else layer2=0.
9. Repeat step8 till j=height and i=width.
10. End of the loop.
11. For j=1and for i=1, find If channel(j,i)>100, Then layer3=channel(j,i) else layer3=0.
12. Repeat step11 till j=height and i=width.
13. End of the loop.
14. For j=1and for i=1, find If channel(j,i)>150, Then layer4=channel(j,i) else layer4=0.
15. Repeat step14 till j=height and i=width.
16. End of the loop.
17. For j=1and for i=1, find If channel(j,i)>200, Then layer5=channel(j,i) else layer5=0.
18. End of the loop.
19. Repeat the same for green channel and blue channel.
20. Initialize block size. Then set the threshold.
21. Initialize the size of the layer to height and width.
22. Assign floor width of a block to number of widths of a block and floor height of a block to number of heights of a block.
23. For j=1 to number of height and for i=1 to number of width, initialize count to zero.
24. For k=start height of a block and for l=start of width of a block, if layer(k,l)==0 then count=count+1.
25. Repeat the step25 till k=end of the start height of a block and l=end of start width of a block.
26. End of the loop.
27. If count>thresh then layer(k,l)=0 else layer(k,l)=255.
28. End of if statement.
29. Initialize starting width =end of the start width +1 and end of start width=end of start width+blocks.
30. Then, if end of the start width >width then end of the start width=width
31. End of if.
32. Initialize starting height= end of the start height+1 and end of the start height=end of height + blocks.
33. Then, if end of the start height >height then end of the start height=height.
34. End of if.
35. Repeat from step17 to 35 till j=height and i=width.
36. End of the loop.

## 3.2. Intersection

The three segmented layers of each channel has to be merged to a single channel. To perform such, we used intersection method. This is done by using formula given below Eq.2.

$$A \cap B = \{x | x \in A \text{ and } x \in B\} \quad (2)$$

The intersection of the segmented layers is as shown in the Fig. 4.

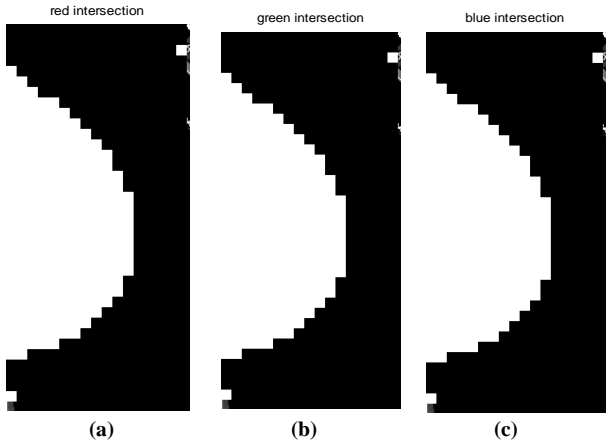

**Fig 4.** Intersection of the segmented layers ( (a) Red segmented channel (b) Green segmented channel (c) Blue segmented channel)

## 3.3. Region of Interest

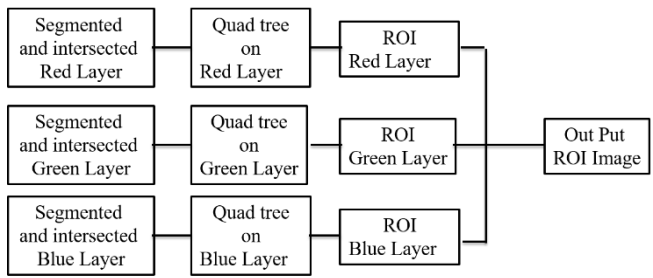

**Fig. 5.** Block diagram of ROI

### 3.3.1. Quad tree

Quad tree is a tree which has four children for each nodes and this method is often used in image compression, also used in calculating image complexity. Quad tree decomposition adopted shows the boundary region of the mass for radiologists to analyze. Divide and conquer rule repeatedly applied divides the image until it meets the criteria. Here, quad tree applied based on entropy divides the area repeatedly which are having high entropy. It ignores the regions having low entropy from division (Eq. 4).

$$\text{Entropy} = -\sum P_i Log_2 P_i \quad (3)$$

The image subjected to quad tree is initially divided into quad blocks, then again subdivided the blocks successively into quadrant until it meets the criteria ie. a thick cluster of edges Fig. 6. The edges arranged closely looks very bright to form thick cluster of bands. The masses size during early stage is about less than 1cm. Our algorithm highlighted those small regions by closer decomposition telling that it needs attention.

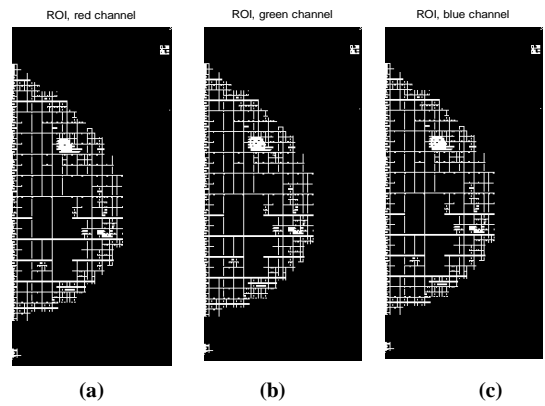

**Fig 6.** Thick cluster of edges ((a) ROI of red channel (b) ROI of green channel (c) ROI of blue channel)

The quad tree outputs of the channels are merged using intersection operations to get the resultant region of interest shown in Fig 7.

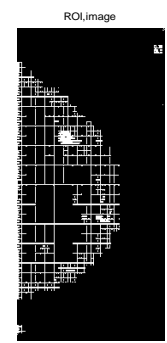

**Fig 7.** Final ROI output

## 4. Results

The automated pinpointing ROI using quad tree decomposition is tested on the Digital Database for Screening Mammography (DDSM) datasets which is a benchmark datasets. These datasets are publicly available which contains both positive and negative sample of images. The informations about suspicious region locations and ground truth for the presence of abnormality in the image are also mentioned in the datasets.

### 4.1. Experimentation on Benchmark Datasets

For the experimentation we took 100 cancer positive images which are already marked by the radiologist and 100 cancer negative images. Since this is a medical image, information loss from the image may lead to misclassification. Hence RGB images are not converted to gray images. Segmentation is performed on each channel separately. We considered different thresholds for layers of each channel which are shown in the table 1.

Table 1. Different thresholds for different layers

| Sl. No. | Red, green, blue layers | Threshold values |
|---|---|---|
| 1 | Layer1 | 0-50 |
| 2 | Layer2 | 50-100 |
| 3 | Layer3 | 100-150 |
| 4 | Layer4 | 150-200 |
| 5 | Layer5 | 200 and above |

As Fig 2 ((b), (c), (d)) depicts layer1 and layer 5 has not provide any information. We performed segmentation on layers 2, 3 and 4 by fixing block size=10 and threshold value=50. All the images persue good segmentation results compared to previous work. The segmented parts are tested by applying quad tree on images containing cancer, which shows the ROI after decomposition. Number of iterations performed in quad tree is 10. It starts from 512, 256, 128, 64, 32, 16, 8, 4, 2, 1. Fig 7(a) shows the area with more division and having more edges which means a group of edges closely arranged to form an abnormal shape are the regions where some abnormality can be predicted. The detection rate is shown about 55 percent on the images.

### 4.2. Comparison Of Normal Images With Cancer Images

Here we compare the cancer positive images with the normal (cancer negative) images to find the differences between them in the results. The most important thing in the cancer image is that the abnormal masses possess high intensity at the center and intensity decreases at the outer side of the boundary. Quad tree supports division when there is a change in the intensities. Hence the quad tree decomposes more at the boundary level of the mass rather than at the center of the mass in the cancer image. Quad tree applied on the normal image shown in Fig. 8.(c) also shows decomposition but here we could not find such divisions since it does not possess masses. This clearly shows that results obtained after applying quad trees for normal and cancer images possessing different way of divisions leading to region of interest in only the cancer images. The region of interest is the place where the radiologists have already marked as cancer and perhaps this again is a cross conformation to give attention on that region. Fig 8 shows the original RGB image, segmented image and ROI image for normal image. Fig 9 shows the original RGB, segmented image and ROI image of cancer image.

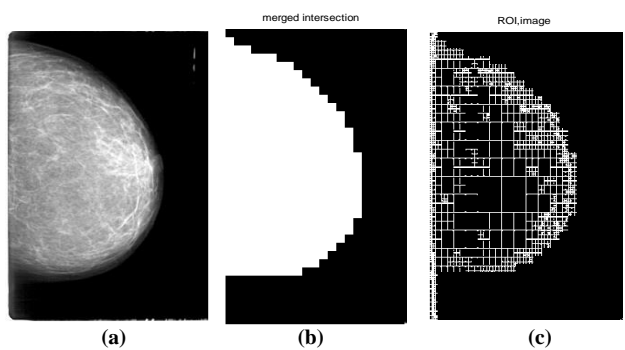

**Fig. 8.** (a) Original image(normal) (b) Segmented image (c) ROI image

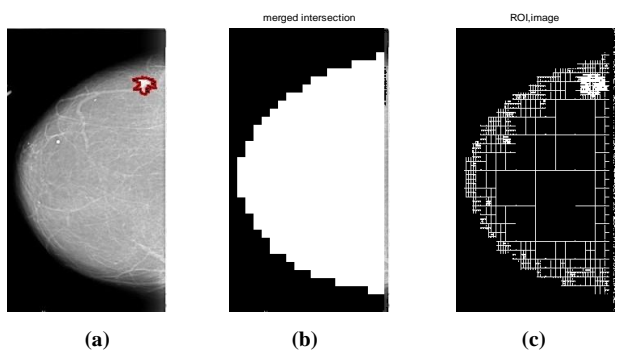

**Fig. 9.** (a) original image(cancer), (b) Segmented image (c) ROI image.

### 4.3. PERFORMANCES

Table 2. Segmentation and location of ROI results

| Image type | Number of images | Properly Segmented images | Percentage of proper segmentation | ROI | Percentage of ROIs |
|---|---|---|---|---|---|
| Cancer images | 200 | 180 | 90% | 120 | 60% |
| Normal images | 100 | 90 | 90% | 80 | 80% |

## 5. Discussion

Our result shown tells that RGB images considered for processing has produced proper segmentation. Sometimes while imaging the breast, pectoral muscle which is a homogeneous triangular part on the top left region will be overlapped with the mass. As per radiologists such overlapped masses on the pectoral muscle are undoubtedly cancer. Hence we did not remove pectoral muscle from the mammograms. Segmentation results produced best accuracy rate. The algorithm designed has given good segmentation for all images but accuracy dropped in locating ROI in images. For images having larger masses, quad tree division gives cluster which is hard to distinguish between normal tissue and abnormal tissue. In future, the thick clusters can be analyzed based on their texture and boundaries for the automated classification of the breast cancer. Extremely dense and heterogeneously dense breasts can also be tested for classification in future.

## 6. Conclusion

The proposed method contributed to improve the segmentation results to the best accuracy for all kinds of breasts without loss of any information. Also suggested that pectoral muscle remove is not necessary as in some images masses might overlap on the pectoral muscle. Quad tree division is a novel approach as it forms a cluster of edges which demands attention for the classification and to find the aggressiveness of that region.